\documentclass[runningheads]{llncs}
\usepackage{graphicx}

\usepackage{bm}
\usepackage{booktabs}
\usepackage{multirow}
\usepackage{bbm}
\usepackage{array}
\usepackage{subfigure}
\usepackage{tikz}
\usepackage{comment}
\usepackage{amsmath,amssymb} 
\usepackage{color}
\usepackage{bibentry}
\usepackage[accsupp]{axessibility}  
\usepackage[marginal]{footmisc}
\usepackage{makecell}
\renewcommand{\thefootnote}

\begin{document}
\pagestyle{headings}
\mainmatter
\def\ECCVSubNumber{1250}  

\title{Towards Calibrated Hyper-Sphere Representation via Distribution Overlap Coefficient for Long-tailed Learning} 

\titlerunning{ Towards Calibrated Hyper-Sphere Representation for Long-tailed Learning}
%

\author{Hualiang Wang\inst{1,3} \thanks{These authors contributed equally.}\and
Siming Fu\inst{1~*}  \and
Xiaoxuan He\inst{1}
\and
Hangxiang Fang\inst{1}  \and
Zuozhu Liu\inst{1,2}  \and
Haoji Hu\inst{1} \thanks{Corresponding author}} 
\authorrunning{H. Wang et al.}
%
\institute{
$^1$College of Information Science and Electronic Engineering, Zhejiang University, China $^2$ZJU-UIUC Institute, Zhejiang University, China $^3$ Angelalign Inc., Shanghai. 
\email{\{hualiang\_wang,fusiming,chuhp,Xiaoxiao\_He,fhx,haoji\_hu\}@zju.edu.cn}\texttt{,}\\
\email{zuozhuliu@intl.zju.edu.cn}
}

\maketitle

\begin{abstract}
 Long-tailed learning aims to tackle the crucial challenge that head classes dominate the training procedure under severe class imbalance in real-world scenarios. However, little attention has been given to how to quantify the dominance severity of head classes in the representation space. Motivated by this, we generalize the cosine-based classifiers to a von Mises-Fisher (vMF) mixture model, denoted as vMF classifier, which enables to quantitatively measure representation quality upon the hyper-sphere space via calculating distribution overlap coefficient. To our knowledge, this is the first work to measure  representation quality of classifiers and features from the perspective of distribution overlap coefficient. On top of it, we formulate the inter-class discrepancy and class-feature consistency loss terms to alleviate the interference among the classifier weights and align features with classifier weights. Furthermore, a novel post-training calibration algorithm is devised to zero-costly boost the performance via inter-class overlap coefficients. Our method outperforms previous work with a large margin and achieves state-of-the-art performance on long-tailed image classification, semantic segmentation, and instance segmentation tasks (e.g., we achieve 55.0\% overall accuracy with ResNetXt-50 in ImageNet-LT). Our code is available at \url{https://github.com/VipaiLab/vMF\_OP.}

\end{abstract}

\section{Introduction}

\label{introduction}

Most real-world data comes with a long-tailed nature: a few head classes contribute the majority of data, while most tail classes comprise relatively few data. An undesired phenomenon is models~\cite{wang2021longtailed,cao2019learning,tan2020equalization} trained with long-tailed data perform better on head classes while exhibiting extremely low accuracy on tail ones.

To remedy it, one of the mainstream insights works on devising balanced classifiers~\cite{2019Decoupling,wusolving,2021Adversarial} against imbalanced data. The cosine-based classifier discards the norms that have been proven to be larger on head classes~\cite{longtailedsurvey}.
The $\tau$-norm classifier~\cite{2019Decoupling} manually shrinks the discrepancy among the norms of classifier weights through a $\tau$-normalization function. In addition, some works~\cite{balanced_loss,LADE,focal,cao2019learning} attach extra margin or scale terms on output scores to prompt classifiers to focus on data-scarce classes. Another prevailing method devotes to learning discriminative features using imbalanced data~\cite{DRO-LT,range_loss,distribution_adaption,cui2021parametric,weng2021unsupervised}. Range loss~\cite{range_loss} is proposed to enlarge the inter-class feature distance and reduce the intra-class feature variation within the mini-batch data. Unsupervised discovery (UD)~\cite{weng2021unsupervised} uses self-supervised learning to help the model highlight tail classes from the feature level. In addition, LDA~\cite{distribution_adaption} transfers the learned feature distribution from the training domain to an ideal balanced domain.

While achieving promising performance, there lack of measures to quantitatively evaluate to what extent these classifiers or features can achieve the presumed ``balanced" classifiers or ``discriminative" features. Hence, one cannot measure how severely head classes dominate the features and classifiers in the high-dimensional representation space, resulting in confusions to guide further optimization for improved long-tailed learning.

To this end, we first extend cosine-based classifier as a von Mises-Fisher (vMF) distribution mixture model on hyper-sphere, denoted as the vMF classifier. Second, based on the representation space constructed by the vMF classifier, we mathematically define a novel measure between two probability density fuctions, denoted as distribution overlap coefficient $o_\Lambda$, to quantify to what extent the classifiers are "balanced" or features are "discriminative". A high $o_\Lambda$ means that the two distributions (classes) are severely intertwined together. We suppose that $o_\Lambda$ among classes in a ``balance'' classifier should be low enough, i.e., one class is not overwhelmingly dominated by other ones. ``Discriminative'' features means  $o_\Lambda$ between features and the corresponding classifier weights is high enough, i.e., features are well matched with correct classes. 

On top of $o_\Lambda$, we provide an explicit optimization objective to boost the representation quality on hyper-sphere, i.e., to allow classifier weights to be distributed separately while aligning the weights of classifiers with features. Specifically, we propose two loss terms: the inter-class discrepancy and class-feature consistency loss. The first one minimizes the overlap among classifier weights, and the second one maximizes the overlap between features and the corresponding classifier weights. To further ease dominance of the head classes in classification decisions during inference, we develop a post-training calibration algorithm for classifier at zero cost based on the learned class-wise overlap coefficients. 

We extensively validate our model on three typical visual recognition tasks, including image classification on  benchmarks (ImageNet-LT~\cite{openlongtailrecognition} and iNaturalist2018~\cite{van2018inaturalist}), semantic segmentation on ADE20K dataset~\cite{zhou2017scene}, and instance segmentation on LVIS-v1.0 dataset~\cite{gupta2019lvis}. The experimental results and ablative study demonstrate our method consistently outperforms the state-of-the-art approaches on all the benchmarks. 

\textbf{Summary of Contributions:}
\begin{itemize}
\item To the best of our acknowledge, we are the first in long-tailed learning to define the distribution overlap coefficient to evaluate representation quality for features and the proposed vMF classifiers.
\item  We formulate overlap-based inter-class discrepancy and class-feature consistency loss terms to alleviate the interference among the classifier weights and align features with classifier weights.
 
\item We develop a post-training calibration algorithm for classifier at zero cost based on the learned class-wise overlap coefficients to ease dominance of the head classes in classification decisions during inference.  

\item Our models outperform previous work with a large margin and achieve state-of-the-art performance on long-tailed image classification, semantic segmentation and instance segmentation tasks.

\end{itemize}

\section{Related works}

\textbf{Classifier design for deep long-tailed learning.} 
In generic visual problems~\cite{learningforrec,momentum}, the common practice of deep learning is to use linear classifier. However, long-tailed class imbalance often results in larger classifier weight norms for head classes than tail classes, which makes the linear classifier easily biased to dominant classes. To address long-tailed class imbalance, researchers design different types of classifiers. Scale-invariant cosine classifier~\cite{2021Adversarial} is proposed, where both the classifier weights and sample features are normalized. The $\tau$-normalized classifier~\cite{2019Decoupling} rectifies the imbalance of decision boundaries by introducing the $\tau$ temperature factor for normalization~\cite{2020Identifying}. Realistic taxonomic classifier (RTC)~\cite{wusolving} addresses the issue with hierarchical classification where different samples are classified adaptively at different hierarchical levels.  GistNet classifier~\cite{liu2021gistnet} leverages the over-fitting to the popular classes to transfer class geometry from popular to few-shot classes. Causal classifier~\cite{tang2020longtailed} records the bias by computing the exponential moving average features during training, and then removes the bad causal effect by subtracting the bias from prediction logits during inference. 
\par
\noindent
\textbf{Representation learning for long-tailed learning.} 
Existing representation learning methods for long-tailed learning mainly focus on  metric learning, prototype learning. 
Metric learning based methods~\cite{kang2021exploring,wang2021contrastive,DRO-LT} explore distance-based losses to learn a more discriminative feature space. LMLE~\cite{LMLE} introduces a quintuple loss to learn representations that maintain both inter-cluster and inter-class margins. Prototype learning based methods~\cite{OLTR,zhu2020inflated} seek to learn class-specific feature prototypes to enhance long-tailed learning performance. 
Open long-tailed recognition (OLTR)~\cite{OLTR} innovatively explores the idea of feature prototypes to handle long-tailed recognition in an open world.  Self-supervised pre-training (SSP)~\cite{yang2020rethinking} uses self-supervised learning for model pre-training, followed by standard training on long-tailed data. 
\par
\noindent
\textbf{von Mises-Fisher Distribution.} In directional statistics, the von Mises–Fisher distribution~\cite{vmf_distribution} is a probability distribution on the hyper-sphere. There are a lot of methods built on von Mises–Fisher distribution in machine learning and deep learning. The vMF Mixture Model (vMFMM)~\cite{hasnat2017mises} proposes SFR model which assumes that the facial features are unit vectors and distributed according to a mixture of vMFs.
The vMF k-means algorithm~\cite{mash2015k} is proposed based on the mixture vMF distribution to unsupervisedly evaluate the compactness and orientation of clusters.  
More recently, the t-vMF similarity\cite{tvmfsimilarity} rebuilds the classifier by the proposed similarity based on vMF distribution to regularize features within deteriorated data.
Sphere Confidence Face~\cite{scf} minimizes KL divergence between spherical Dirac delta and $r$-radius vMF to achieve superior performance on face uncertainty learning.

Different from all them, to our best acknowledge, we are the first to quantify the distribution overlap coefficient between vMF distributions.
Benefiting from it, we conduct a series of comprehensive and in-depth analyses to explore how to achieve high-quality representation space built upon vMF distribution.

\begin{figure}[t]
  \centering
  \includegraphics[width=1\textwidth]{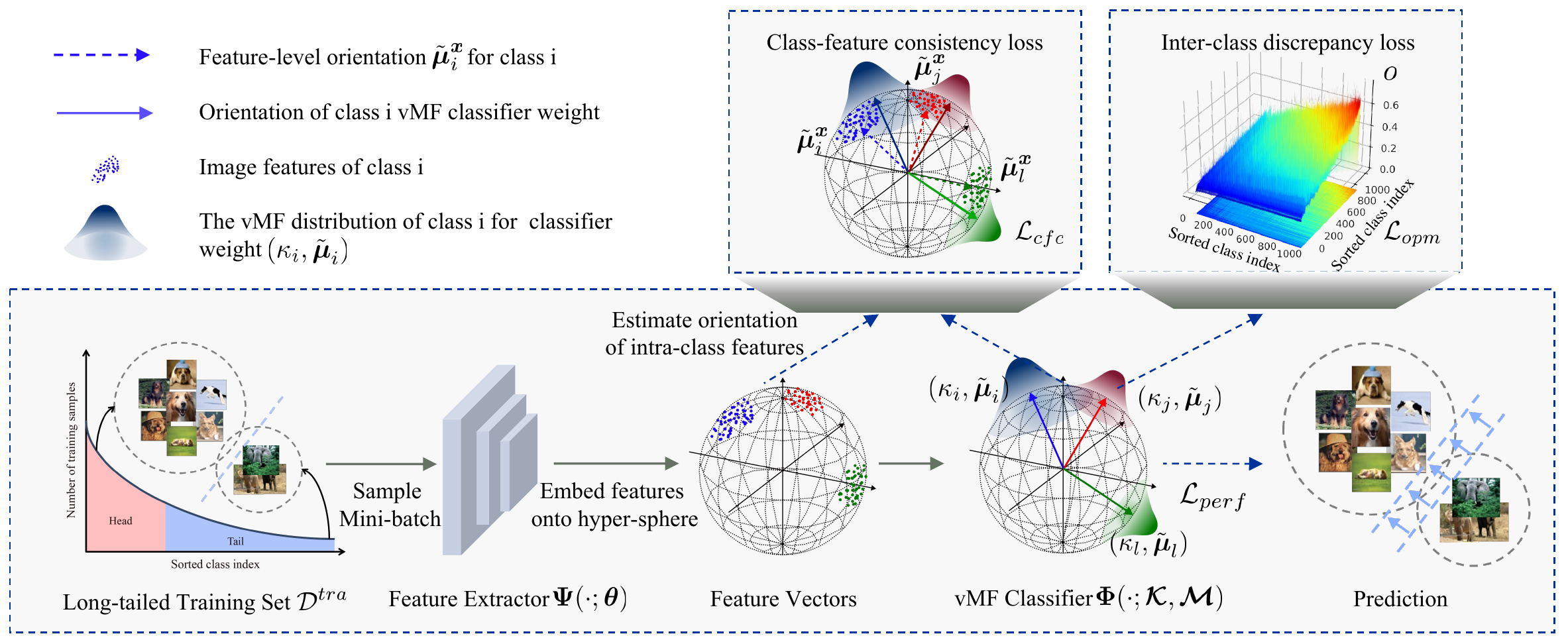}  
  \caption{ Overview of our proposed method during the training period. \textbf{Bottom box} consists of the following steps in sequence: sampling a mini-batch images $\mathcal{B}$ from training set $\mathcal{D}^{tra}$, learning features by the feature extractor $\bm{\Psi}(\cdot; \bm{ \theta })$, embedding features onto hyper-sphere, predicting output via our proposed vMF classifier $ \bm{ \Phi }( \cdot ; \bm{ \mathcal{K} }, \bm{ \mathcal{M} } )$ and calculating the performance loss value.
  \textbf{Upper boxes} introduce our proposed the class-feature consistency loss term $\mathcal{L}_{cfc}$ and inter-class discrepancy loss term $\mathcal{L}_{icd}$.
      }
	\label{fig:pipline}
\end{figure}

\section{Methodology}
First, we briefly review the canonical pipeline of long-tailed learning, exemplified by long-tailed image classification, and elaborate on our proposed vMF classifier. 
Afterward, we mathematically define the distribution overlap coefficient. On top of it, we further present the proposed the inter-class discrepancy loss and class-feature consistency loss terms. 
Finally, a post-training calibration algorithm is devised to zero-costly boost performance.
\subsection{ Build vMF Classifier on Hyper-Sphere}
\label{31}
Let $ \mathcal{ D }^{tra} = \{ \bm{I}^l, y^l\}$, $ l\in\{1, \cdots, N\} $ be the training set, where $\bm{I}^l$ denotes an image sample and $y^l =i$ indicates it belongs to class $i$. 
Let $C$ be the total numbers of classes, $n_i$ be the number of samples in class $i$, where $\sum_{i=1}^{C} n_{i}=N$. 
The class prior distribution on training set can be defined as $p^{tra}_{\mathcal{D}}(i) = n_i / N$.

As shown in Fig.~\ref{fig:pipline}, given a pair $ (\bm{I}^l, y^l) $ sampled from a mini-batch $ \mathcal{ B } \subset \mathcal{ D }^{tra} $, feature vector $\bm{x}^l = \bm{\Psi}(\bm{I}^l; \bm{ \theta }) \in \mathbb{R}^{1 \times d}$ is extracted by the feature extractor $ \bm{ \Psi }(\cdot; \bm{ \theta }) $, of which learnable parameter $\bm{\theta}$ is instantiated by a neural network (e.g., ResNet).
Then $\bm{x}^l$ is projected onto the unit hyper-sphere $\mathbb{S}^{d-1}$ via $\tilde{ \bm{x} }^l = \bm{x}^l / \Vert \bm{x}^l \Vert_2$  and subsequently fed into the vMF classifier.

We depict the classifier with $C$ classes as a mixture model with $C$ von Mises-Fisher distributions on $\mathbb{S}^{d-1}$, each class containing two variables: the compactness $\kappa_i \in \mathbb{R}^+$ and the unit orientation vector  $\tilde{ \bm{\mu} }_i \in \mathbb{R}^{1 \times d}$.
Consequently,  vMF classifier is well-defined as $ \bm{ \Phi }( \cdot ; \bm{ \mathcal{K} }, \bm{ \mathcal{M} } )$,
where $\bm{ \mathcal{K} } = \{ \kappa_1, ..., \kappa_C \}$ and $\bm{ \mathcal{M} } = \{ \tilde{ \bm{\mu} }_1, ..., \tilde{ \bm{\mu} }_C\}$ are learnable compactness and orientation vectors for $C$ classes, respectively.
The probability density function (PDF) $p(\tilde{ \bm{x} } | \kappa_i, \tilde{ \bm{ \mu } }_i )$ of $i$-th class is mathematically defined as:
\begin{equation}
\begin{aligned}
  \label{pdf}
  p(\tilde{ \bm{x} } | \kappa_i, \tilde{ \bm{ \mu } }_i ) = C_d(\kappa_i) e^{ \kappa_i \cdot \tilde{\bm{x}} \tilde{ \bm{\mu} }_i^{\top} } = \frac{ {\kappa_i}^{ \frac{d}{2}-1} }{ (2\pi)^{ 
    \frac{d}{2}} \cdot I_{ \frac{d}{2}-1}(\kappa_i) } e^{ \kappa_i \cdot \tilde{\bm{x}} \tilde{ \bm{\mu} }_i^{\top} },
\end{aligned}
\end{equation}
where $I_v(\kappa)$ is the modified Bessel function~\cite{bessel} of the first kind of real order $v$ and $C_d(\kappa)$ is a normalization constant. 

From the view of Bayes Theorem~\cite{bayes}, given the class prior distribution $p^{tra}_{\mathcal{D}}(i)$ and $p(\tilde{ \bm{x} }^l | \kappa_i, \tilde{ \bm{ \mu } }_i )$, 
the probability $p^l_i $ for  $\bm{I}^l$ belonging to class $i$ can be formulated by the posterior probability $p(y^l=i | \tilde{ \bm{x} }^l)$ as:
\begin{equation}
  \begin{aligned}
    \label{poster}
    p^l_i = p(y^l=i | \tilde{ \bm{x} }^l) = \frac{ p^{tra}_{\mathcal{D}}(i) \cdot p(\tilde{ \bm{x} }^l | \kappa_i, \tilde{ \bm{ \mu } }_i )    }{ \sum_{j=1}^C p_{\mathcal{D}}^{tra}(j) \cdot p(\tilde{\bm{x}}^l | \kappa_j, \tilde{\bm{ \mu }}_j )  }.
  \end{aligned}
\end{equation}
 
Eq.~\ref{poster} is the formulation of our vMF classifier. 
Our vMF classifer degrades to a balanced cosine classifier~\cite{balanced_loss} with a temperature $\sigma$, when $\kappa_i = const~ \sigma, \forall i \in [1, C] $. 

The performance loss $\mathcal{L}_{perf}$ of the mini-batch $\mathcal{B}$ is calculated by the cross-entropy function as follows:
\begin{align}
  \label{loss_perform}
  \mathcal{L}_{perf} = -\frac{1}{N'} \sum_{l=1}^{N'} \sum_{i=1}^C \mathbbm{1}[y^l=i] \cdot \log{p^l_i},
\end{align}
where $\mathbbm{1}[y=i]$ is the binary indicator that denotes whether the corresponding image comes from the $i$-th class and $N'$ is the number of samples in a mini-batch. 

The total loss $\mathcal{L}$ for mini-batch $\mathcal{B}$ in one iteration is calculated as:
\begin{equation}
  \begin{aligned}
  \label{total_loss}
\mathcal{L} = \mathcal{L}_{perf} + \lambda \cdot ( \mathcal{L}_{icd} + \mathcal{L}_{cfc} ), 
\end{aligned}
\end{equation}
where $\mathcal{L}_{icd}$ and $\mathcal{L}_{cfc}$ are proposed additional loss terms to regularize feature and classifier, which will be introduced in the subsequent subsection.
$\lambda$ is a hyper-parameter to adjust the weight of additional loss terms.

\setlength{\tabcolsep}{4pt}
\begin{table}[t]
  \centering
  \caption{Derivatives for compactness and orientation of vMF classifier.}
  \label{derivation_table}
  \begin{tabular}{p{1.cm}<{\centering}|p{4.75cm}<{\centering}|p{4.75cm}<{\centering}}
  \toprule
       & $ \partial o_\Lambda $ & $ \partial \log p_i^l$ \\ 
  \midrule
  $\partial \kappa_i$ &  $ o^2_\Lambda \cdot \frac{ \partial A_d(\kappa_i)}{ \partial \kappa_i} \cdot (  \kappa_j \cdot \tilde{ \bm{\mu} }_i \tilde{ \bm{\mu} }_j^\top - \kappa_i ) $  &  $ ( 1 - p_i^l ) \cdot ( \tilde{\bm{x}}^l \tilde{ \bm{\mu} }_i^{\top} - A_d(\kappa_i) )  $  \\[3.5pt] 
  
  $ \partial \kappa_j$ & $ o_\Lambda^2 \cdot ( A_d(\kappa_i) \cdot \tilde{ \bm{\mu} }_i \tilde{ \bm{\mu} }_j^\top - A_d(\kappa_j) )$  &  $  - p_j^l  \cdot ( \tilde{\bm{x}}^l \tilde{ \bm{\mu} }_j^{\top} - A_d(\kappa_j) )  $  \\[3.5pt] 
  
  $ \partial \tilde{ \bm{ \mu } }_i $ &  $o_\Lambda^2 \cdot \kappa_j \cdot A_d(\kappa_i) \cdot \tilde{ \bm{ \mu } }_j$   &  $ ( 1 - p_i^l ) \cdot \kappa_i \cdot \tilde{\bm{x}}  $  \\[3.5pt] 
  
  $ \partial \tilde{ \bm{ \mu } }_j $ & $o_\Lambda^2 \cdot \kappa_j \cdot A_d(\kappa_i) \cdot \tilde{ \bm{ \mu } }_i$ &  $  - p_j^l  \cdot \kappa_j \cdot  \tilde{\bm{x}} $  \\[3.5pt]
  \bottomrule
  \end{tabular}
  \end{table}

\subsection{Quantify Distribution Overlap Coefficient on Hyper-Sphere}
\label{32}
As aforementioned, we geometrically depict the classifier as a set of vMF distributions on $\mathbb{S}^{d-1}$. 
The distribution overlap coefficient~\cite{overlap} is mathematically explained as the area of intersection between two probability density functions. 
Based on it, we mathematically quantify distribution overlap coefficient to measure the intersection degree of two classes (vMF distribution) in the $\mathcal{S}^{d-1}$. 
In this paper, we provide the analytic expression $o_\Lambda$ based on Kullback-Leibler divergence~\cite{kl} for the vMF distribution~\cite{diethe2015note}. Specifically, $o_\Lambda$ is defined as:
\begin{equation}
\begin{aligned}
  o_\Lambda(\kappa_i, \kappa_j, \tilde{ \bm{\mu} }_i,  \tilde{ \bm{\mu} }_j) = \frac{1}{ 1 + KL\{ p( \tilde{ \bm{x} } | \kappa_i, \tilde{ \bm{\mu} }_i),  p( \tilde{ \bm{x} } | \kappa_j, \tilde{ \bm{\mu} }_j) \}  },
\end{aligned}
\label{overlap1}
\end{equation}
where $KL\{ p( \tilde{ \bm{x} } | \kappa_i, \tilde{ \bm{\mu} }_i),  p( \tilde{ \bm{x} } | \kappa_j, \tilde{ \bm{\mu} }_j) \} $ is the Kullback-Leibler divergence between two vMF distributions, abbreviated as $KL_{ij}$:
\begin{equation}
  \begin{aligned}
  KL_{ij} &= - \int_{ \tilde{\bm{x}} } p( \tilde{ \bm{x} } | \kappa_i, \tilde{ \bm{\mu} }_i) \cdot  \ln \frac{ p( \tilde{ \bm{x} } | \kappa_j, \tilde{ \bm{\mu} }_j) }{ p( \tilde{ \bm{x} } | \kappa_i, \tilde{ \bm{\mu} }_i) }  \,d\tilde{\bm{x}} \\
   &= \ln \frac{C_d(\kappa_i)}{C_d(\kappa_j)} \cdot \underbrace{ \int_{ \tilde{\bm{x}} } C_d(\kappa_i) \cdot e^{ \kappa_i \cdot \tilde{\bm{x}} \tilde{ \bm{\mu} }_i^{\top} } \,d\tilde{\bm{x}} }_{=1}\\
   &+ \underbrace{ ( \int_{ \tilde{\bm{x}} } \tilde{\bm{x}} \cdot C_d(\kappa_i) \cdot e^{ \kappa_i \cdot \tilde{\bm{x}} \tilde{ \bm{\mu} }_i^{\top} }  \,d\tilde{\bm{x}}  ) }_{= \mathbb{E}[ \tilde{\bm{x}} ] = A_d(\kappa_i) \cdot \tilde{\bm{\mu}}_i } ( \kappa_i \cdot \tilde{\bm{\mu}}_i^\top - \kappa_j \cdot \tilde{\bm{\mu}}_j^\top  ) \\
   &= \ln \frac{C_d(\kappa_i)}{C_d(\kappa_j)} + A_d(\kappa_i) \cdot ( \kappa_i - \kappa_j \tilde{\bm{\mu}}_i \tilde{\bm{\mu}}_j^{\top}),
\end{aligned}
\label{eqn:KL}
\end{equation}
where $A_d(\kappa_i) = I_{d/2}(\kappa_i) / I_{d/2 - 1}(\kappa_i) $ is non-decreasing and $0  A_d(\kappa_i)  1 $. $ \mathbb{E}[ \tilde{\bm{x}} ] $ is the expectation vector for $\tilde{\bm{x}} \sim p( \tilde{ \bm{x} } | \kappa_i, \tilde{ \bm{\mu} }_i) $~\cite{new_vmf}. Generally $0 o_\Lambda \leq 1$. $o_\Lambda = 1$ (i.e., $\kappa_i = \kappa_j$ and $\tilde{\bm{\mu}}_i \tilde{\bm{\mu}}_j^{\top}=1$) means they are completely congruent. $o_\Lambda \rightarrow 0$ indicates there is nearly no intersection between two distributions.

\begin{figure}[t!]
  \centering
  \subfigure[$o_{\Lambda}( \kappa_i, \kappa_j, \tilde{ \bm{\mu} }_i, \tilde{ \bm{\mu} }_j )$]{ 
  \begin{minipage}[t]{0.3\textwidth}
  \centering   
  \includegraphics[width=1\textwidth]{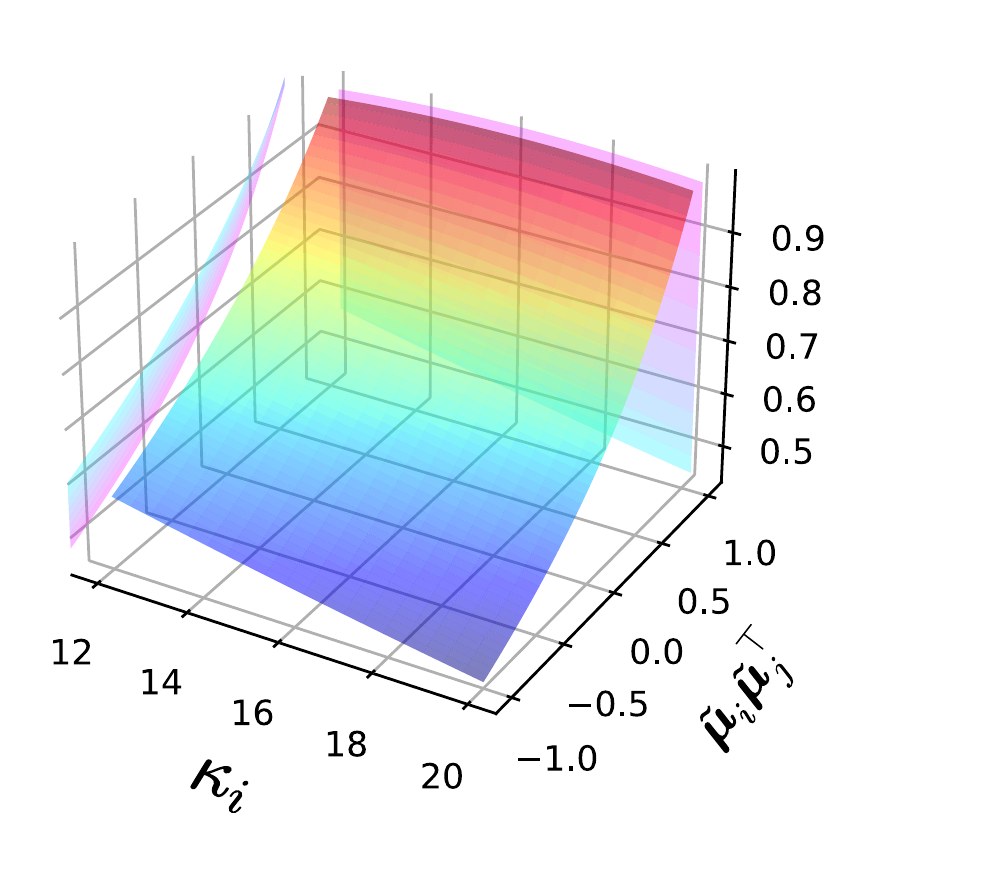} 
  \end{minipage}
  }
  \subfigure[$\frac{\partial o_{\Lambda}(\kappa_i, \kappa_j, \tilde{\bm{\mu}}_i, \tilde{\bm{\mu}}_j)}{ \partial \kappa_i }$]{   
  \begin{minipage}[t]{0.3\textwidth}
  \centering
  \includegraphics[width=1\textwidth]{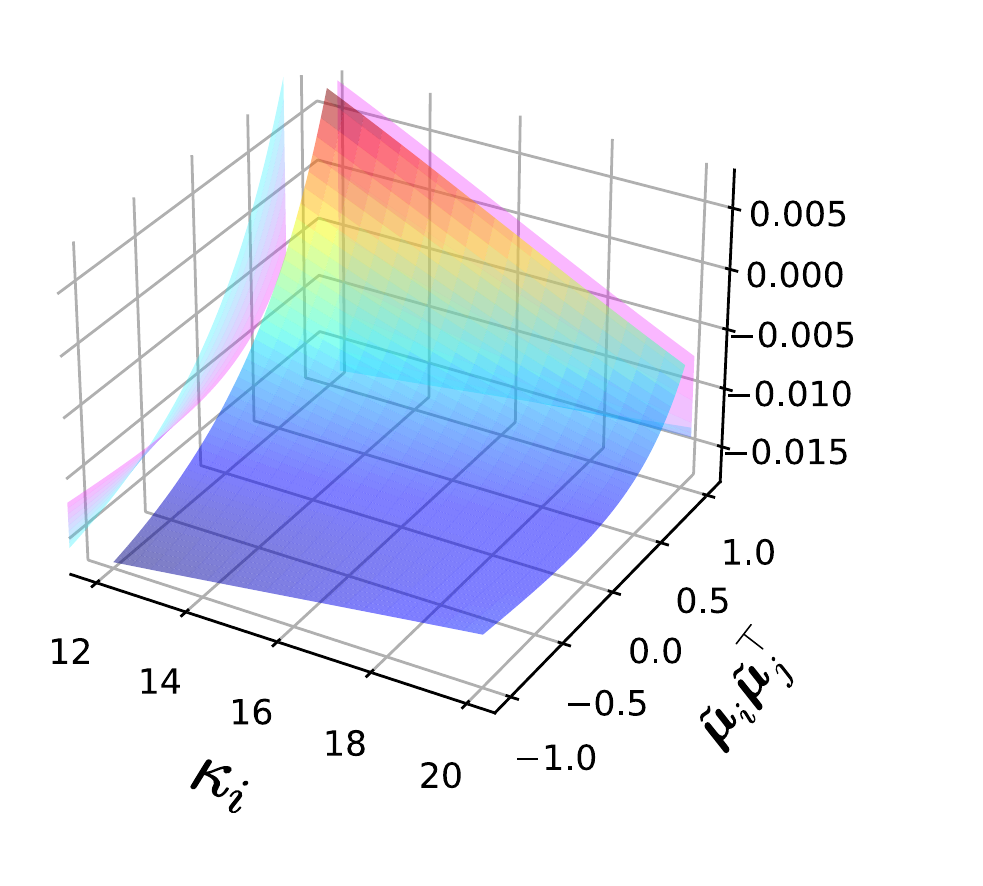}  
  \end{minipage}
  }
  \subfigure[ $\frac{\partial o_{\Lambda}(\kappa_i, \kappa_j, \tilde{\bm{\mu}}_i, \tilde{\bm{\mu}}_j)}{ \partial \tilde{ \bm{\mu} }_i \tilde{ \bm{\mu} }_j^\top }$ ]{   
  \begin{minipage}[t]{0.3\textwidth}
  \centering    
  \includegraphics[width=1\textwidth]{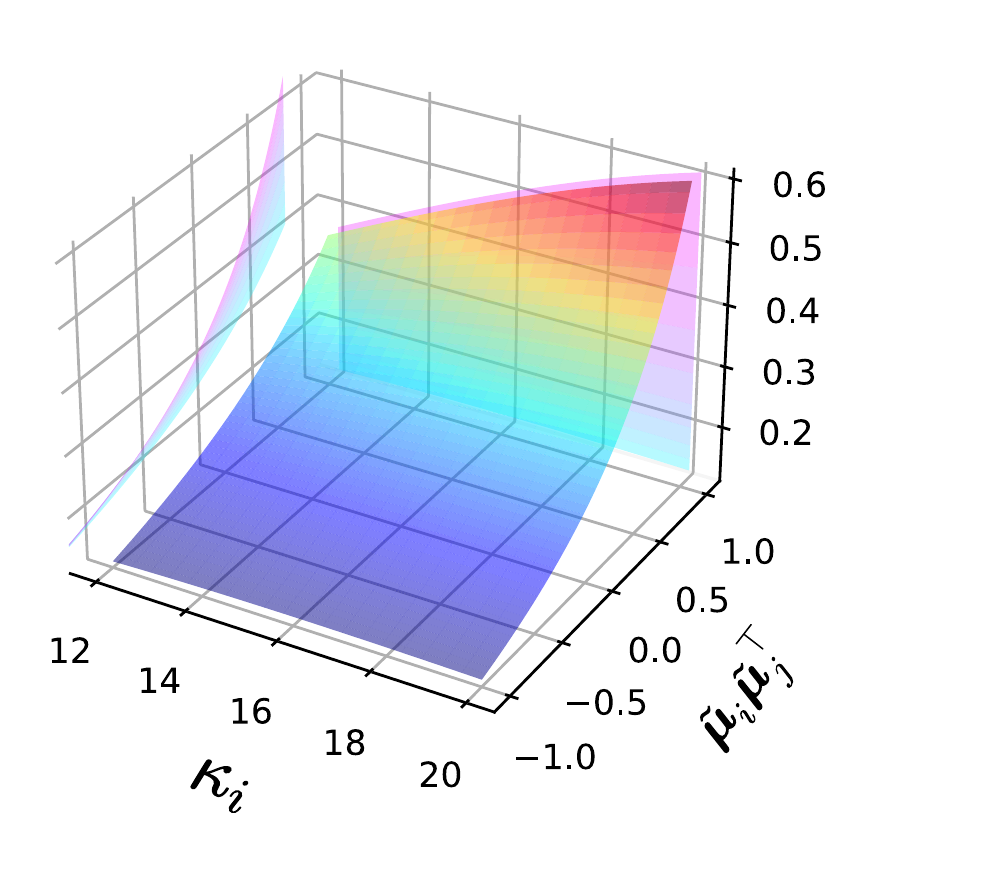}  
  \end{minipage}
  }
  \caption{Visualization of overlap coefficient $o_\Lambda(\kappa_i, \kappa_j, \tilde{\bm{\mu}}_i, \tilde{\bm{\mu}}_j)$ and partial derivatives for $\kappa_i$ and $\tilde{ \bm{\mu} }_i \tilde{ \bm{\mu} }_j^\top$. To exhibit them in 3D coordination, $\kappa_j$ is fixed to a certain value, instantiated as $16$. $\kappa_i$ and $\tilde{ \bm{\mu} }_i \tilde{ \bm{\mu} }_j^\top$ ($\tilde{\bm{\mu}}_i \in \mathbb{R}^{1 \times 512}$ ) are uniformly sampled 100 values from range [12, 20] and range [-1, 1], respectively.  }
  \label{overlap}
\end{figure}

The derivatives of $\kappa_i$, $\kappa_j$, $\tilde{ \bm{\mu} }_i$ and $\tilde{ \bm{\mu} }_j$ for $o_\Lambda$ are listed as the Col 1 of Tab.~\ref{derivation_table}.
And visualization for them is demonstrated in Fig.~\ref{overlap}. Specifically, the partial derivative with respect to $\tilde{\bm{\mu}}_i \tilde{\bm{\mu}}_j^{\top}$ is non-negative.

The partial derivatives with respect to $\kappa_i$ or $\kappa_j$ are non-monotonous. An empirical conclusion is that $\kappa_i$ and $\kappa_j$ need to be kept at the same order of magnitude to achieve guaranteed performance, when using $o_\Lambda$ as the optimization objective.

\subsection{Improve Representation of Feature and Classifier via $o_\Lambda$}
\subsubsection{Inter-class Discrepancy Loss.}
To achieve the discriminative representation space in long-tailed learning, we seek to optimize our vMF classifier via shrinking the overlap among classes as much as possible to alleviate the overwhelm of the head classes on the tail ones.
We denote the above optimization objective as the inter-class discrepancy loss term $\mathcal{L}_{icd}$, which acts function on the weights $\bm{\mathcal{K}}$ and $\bm{\mathcal{M}}$ of the vMF classifier.

First, we measure the average overlap coefficient $o_i$ among class $i$ and all the other classes, formulated by:
\begin{equation}
  \begin{aligned}
    o_i &= \frac{1}{C-1} \sum_{j=1, j \ne i}^C o_\Lambda(\kappa_i, \kappa_j, \tilde{\bm{\mu}}_i, \tilde{\bm{\mu}}_j).
    \label{avg_op}
  \end{aligned}
\end{equation}

Furthermore, we define the inter-class discrepancy loss term $\mathcal{L}_{icd}$ as:
\begin{equation}
  \begin{aligned}
    \label{icd}
    \mathcal{L}_{icd} &= \frac{1}{C} \sum_{i=1}^C o_i,
  \end{aligned}
\end{equation}
 
 The proposed $\mathcal{L}_{icd}$ minimizes the average distribution overlap coefficient to regularize distributions, contributing to a more distinction-prone classifier on $\mathbb{S}^{d-1}$.

\subsubsection{Class-Feature Consistency Loss.} 
In addition, the poorly matching between the feature vectors and the corresponding classifier weights derives unsatisfied performance, especially for the sample-starved classes.
Class-feature consistency loss term $\mathcal{L}_{cfc}$ is proposed to alleviate the above issue by aligning features with the corresponding classifier weights as far as possible. 

Specifically, we first fit the class-wise feature distribution ($\kappa^{\bm{x}}$, $\tilde{ \bm{\mu} }^{\bm{x}}$) within the mini-batch $\mathcal{B}$. The class set involved in $\mathcal{B}$ is denote as $\mathcal{C}'$.
For a certain class $i \in \mathcal{C}'$, the feature-level orientation vector $\tilde{ \bm{\mu} }_i^{\bm{x}}$ is defined as:
\begin{equation}
  \begin{aligned}
    \label{mu_x}
    \tilde{ \bm{\mu} }_i^{\bm{x}} =  \frac{ \sum_{l=1, y^l = i}^{N'} \bm{x}^l  }{ \Vert \sum_{l=1, y^l = i}^{N'} \bm{x}^l  \Vert_2}.
  \end{aligned}
\end{equation}

Considering that the compactness $\kappa$ is over-sensitive to sample number and intractable to be estimated~\cite{hasnat2017mises}, $\kappa$ is shared between the feature and the corresponding classifier weight, i.e., feature-level compactness $\kappa^{\bm{x}}_{i}$ for class $i$ is equal to $\kappa_i$.
Then, $\mathcal{L}_{cfc}$ is formulated as following:
\begin{equation}
  \begin{aligned}
    \label{cfc}
    \mathcal{L}_{cfc} = \mathbb{E}_{i \in \mathcal{C}^{'}}[1 - o_\Lambda(\kappa_i, \kappa_{i}^{\bm{x}}, \tilde{\bm{\mu}}_i, \tilde{\bm{\mu}}_i^{\bm{x}})  ],
  \end{aligned}
\end{equation}
where $\mathbb{E}$ indicates the average function. $\mathcal{L}_{cfc}$ is, in effect, equivalent to maximizing the distribution overlap coefficient between features and the corresponding classifier weights.

\subsection{Calibrate Classifier Weight beyond Training via $o_\Lambda$}
\label{sec:calibrate_classifier}
\begin{figure}[t]
  \centering
  \includegraphics[width=1\textwidth]{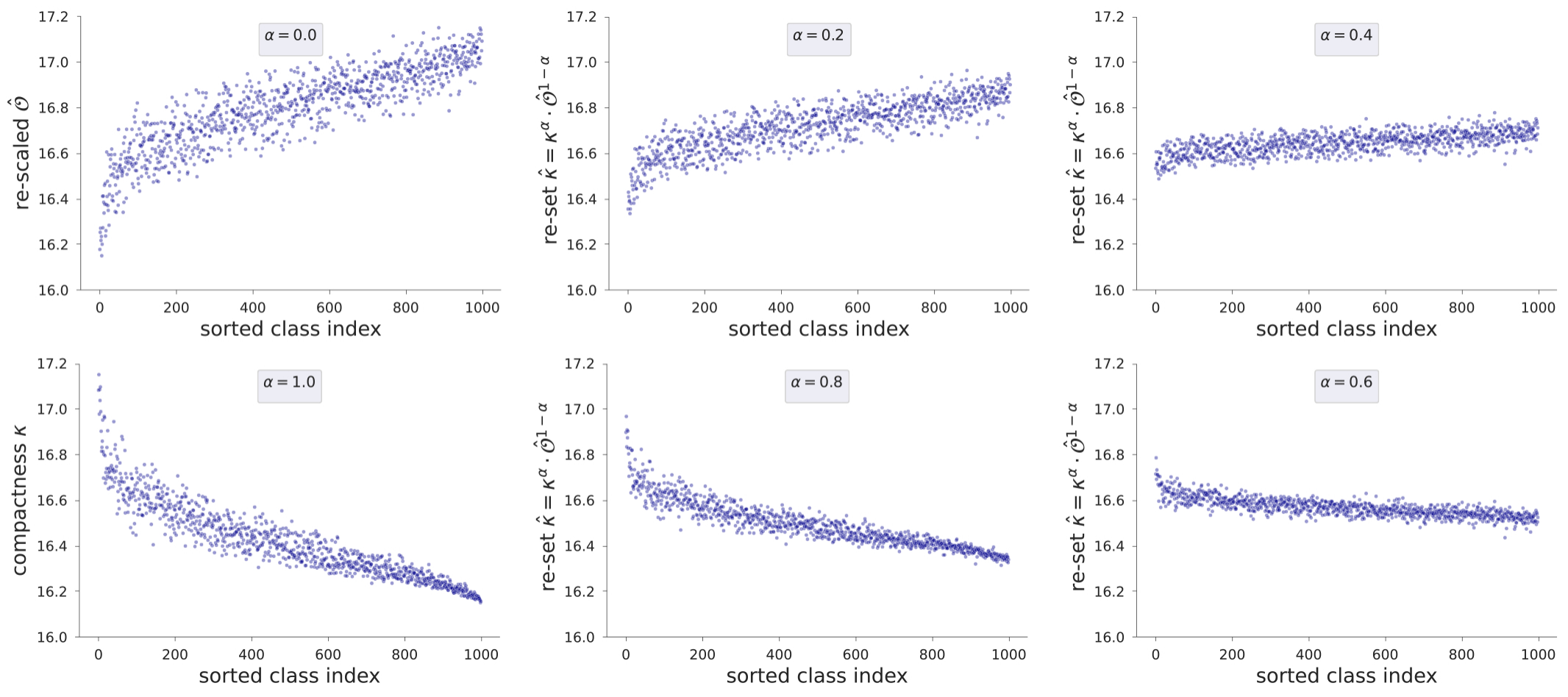}  
  \caption{ The calibrated compactness of vMF classifier (trained on ImageNet-LT
with ResNetXt-50 feature extractor). Under different $\alpha$ settings, we adjust $\kappa$ via Eq.~\ref{norm} and Eq.~\ref{reset}. Each picture represents the value of re-scaled $\hat{ \bm{ \mathcal{K} } }$ when $\alpha$ equals to the corresponding value. When $\alpha = 0$, it indicates $\hat{\kappa}_i = \hat{ o }_i$, while $\alpha = 1$ , $\hat{\kappa}_i = \kappa_i$. }
  \label{fig:calibrate kappa}
\end{figure}

Despite exerting additional loss terms to regularize features and classifiers, the overwhelm of the head classes on the tail ones is, in effect, tough to eradicate under a highly imbalanced dataset.
We visualize the compactness of the classifier and the average overlap coefficients from a well-trained model, as demonstrated in Col 1 of Fig.~\ref{fig:calibrate kappa}.
The head classes share larger compactness and smaller overlap coefficients, however, the case for tail ones is reversed.

A general summary of the calibration strategy is that increase the compactness for classes that are severely overlapped with other classes.
Specifically, given a well-trained vMF classifier $ \bm{ \Phi }( \cdot ; \bm{ \mathcal{K} }, \bm{ \mathcal{M} } )$, we first apply Eq.~\ref{avg_op} to obtain the average overlap coefficient for each class, denoted as $\bm{ \mathcal{O} }= \{o_1, ..., o_C \}$.
Then we use a maximum-minimum normalization strategy to reconcile $\bm{ \mathcal{O} } $ to the same value range as $\bm{ \mathcal{K} }$, to make sure that both are on the same order of magnitude by:
\begin{equation}
  \begin{aligned}
    \label{norm}
    \hat{ o }_i = \frac{ o_i - o^{min} }{ o^{max} - o^{min} } \cdot ( \kappa^{max} - \kappa^{min} ) + \kappa^{min},
  \end{aligned}
\end{equation}
where $o^{max}$ and $o^{min}$ are maximum and minimum values of set $\bm{ \mathcal{O} }$, respectively, as well as $\kappa^{max}$ and $\kappa^{min}$. We reset compactness vector as $\hat{ \bm{ \mathcal{K} } } = \{ \hat{\kappa}_1, ..., \hat{\kappa}_C \}$, formulated as following:
\begin{equation}
  \begin{aligned}
    \label{reset}
    \hat{\kappa}_i = \kappa_i^\alpha \cdot \hat{ o }_i^{ 1 - \alpha},
  \end{aligned}
\end{equation}
 $\alpha \in [0, 1]$ is a hyper-parameter to balance the importance contribution to the re-scaled $\hat{ \bm{ \mathcal{K} } }$ as shown in Fig.~\ref{fig:calibrate kappa}. 
 In the inference period, we comply with a canonical assumption that the classes on the test set follow the uniform distribution, i.e., $ p_{\mathcal{D}}^{test}(i) = 1 / C $. Consequently, we replace $p^{tra}_{\mathcal{D}}(i)$ by $p^{test}_{\mathcal{D}}(i)$
in Eq.~\ref{poster}, and the vMF classifier is calibrated as $ \bm{ \Phi }( \cdot ; \hat{\bm{ \mathcal{K} }}, \bm{ \mathcal{M} } )$. 

Moreover, our post-training calibration algorithm is capable of extending to several wide-used classifiers for cost-free performance boosting. 
Next, we instantiate how to apply the algorithm above to calibrate the weights of $\tau$-norm~\cite{2019Decoupling}, causal classifiers~\cite{tang2020long} and linear classifiers.
Given the weight vector $\bm{w^{\tau}_i}$ of class $i$ from a well-trained $\tau$-norm classifier $\bm{W}^{\tau}$, 
we equivalently convert $\bm{w^{\tau}_i}$ into compactness $\kappa_i = \Vert \bm{w^{\tau}_i} \Vert_2^{1 - \tau}$ and orientation vector $\tilde{\bm{\mu}}_i = \bm{w^{\tau}_i} / \Vert \bm{w^{\tau}_i} \Vert_2$.
After calibration via Eq.~\ref{norm} and Eq.~\ref{reset}, $\bm{w^{\tau}_i}$ is rebuilt by producting orientation vector and re-balanced compactness together.
Along the same lines, the weight vector $\bm{w^{cau}_i}$ for a well-trained causal classifier $\bm{W}^{cau}$ is converted to $\kappa_i = \Vert \bm{w^{cau}_i} \Vert_2/ (\Vert \bm{w^{cau}_i} \Vert_2 + \gamma)$ and $\tilde{\bm{\mu}}_i = \bm{w^{cau}_i} / \Vert \bm{w^{cau}_i} \Vert_2$. The weight vector $\bm{w^{lin}_i}$ for a well-trained linear classifier $\bm{W}^{lin}$ is converted to $\kappa_i = \Vert \bm{w^{lin}_i} \Vert_2 $ and $\tilde{\bm{\mu}}_i = \bm{w^{lin}_i} / \Vert \bm{w^{lin}_i} \Vert_2$.
$\gamma$ and $\tau$ are both the hyper-parameters for classifiers above. (Detail proofs in Appendix $\textbf{A.2}$)


\section{ Experiments}
In this section, we conduct a series of experiments to validate the effectiveness of our method. Below we present our experimental analysis  in Sec.~\ref{section:Long-tailed Image Classification Task}, followed by our results on semantic segmentation task and instance segmentation task in Sec.~\ref{section:Semantic Semgnetaion on ADE20k Dataset} and ablation study in Sec.~\ref{section:ablation study}.

\setlength{\tabcolsep}{4pt}
\begin{table}[t]
\centering
\caption{Results on ImageNet-LT in terms of accuracy (Acc) under 90 and 200 training epochs. In
this table, CR, DT,  RL and CD indicate class re-balancing, decouple training, representation learning and classifier design, respectively. $\dagger$ indicates only vMF classifier is applied w/o additional loss terms and post-training calibration algorithm. }
\begin{tabular}{l|l|ccccccccc}
\toprule
\multirow{2}{*}{Type} & \multirow{2}{*}{Method} & \multicolumn{4}{c}{90 epochs}                      &  & \multicolumn{4}{c}{200 epochs} \\ \cmidrule{3-6} \cmidrule{8-11} 
                      &                         & Many & Med. & Few & All &  & Many  & Med. & Few & All \\ \midrule
Baseline            & Softmax               & 66.5 & 39.0   & 8.6  & 45.5   & &66.9 & 40.4 &12.6 &46.8\\ \midrule
\multirow{5}{*}{CR} & Focal Loss~\cite{focal}              & 66.9 & 39.2   & 9.2  & 45.8  & &67.0 & 41.0 &13.1 &47.2 \\
                    & BALMS~\cite{balanced_loss}      & 61.7 & 48.0   & 29.9 & 50.8  & &62.4 &47.7 &32.1 &51.2 \\
                    & LDAM~\cite{cao2019learning}    & 62.3 & 47.4   & 32.5 & 51.1  & &60.0 &49.2 &31.9 & 51.1 \\
                    & LADE~\cite{LADE}                   & 62.2 & 48.6   & 31.8 & 51.5 & &63.1 &47.7 &32.7 & 51.6 \\ 
                    &DisAlign~\cite{zhang2021distribution}                     &62.7  &52.1    &31.4  & 53.4  & &-&-&-&-  \\
                     \midrule
\multirow{4}{*}{DT} &IB-CRT~\cite{2019Decoupling}    & 62.6 & 46.2   & 26.7 & 49.9& &64.2 &46.1 &26.0 &50.3  \\
                    &CB-CRT~\cite{2019Decoupling}    &62.4 &39.3 &14.9&44.9 & &60.9 &36.9 &13.5 &43.0\\
                    & MiSLAS~\cite{mislas}              & 62.1 & 48.9   & 31.6 & 51.4  && 65.3 & 50.6 &33.0 & 53.4   \\ 
                    &xERM$_{\rm{TDE}}$~\cite{xERM} &-&-&-&-& &68.6&50.0 &27.5 &54.1\\\midrule
\multirow{4}{*}{RL} & OLTR~\cite{OLTR}                 & 58.2 & 45.5   & 19.5 & 46.7 & &62.9 &44.6 &18.8 &48.0  \\
                    & SSP~\cite{yang2020rethinking}                   & 65.6 & 49.6   & 30.3 & 53.1  & &67.3 & 49.1 &28.3 &53.3 \\
                    &DRO-LT~\cite{DRO-LT} &-&-&-&-& &64.0 & 49.8 &33.1 &53.5\\
                    & PaCo~\cite{cui2021parametric}                  & 59.7 & 51.7   & 36.6 & 52.7  & &63.2 &51.6 &39.2 &54.4\\ \midrule
\multirow{2}{*}{CD} & $\tau$-norm~\cite{2019Decoupling}         & 61.8 & 46.2   & 27.4 & 49.6   & &-&-&-&-\\
                    & TDE~\cite{tang2020long}               & 63.0 & 48.5   & 31.4 & 51.8   & &64.9 &46.9 &28.1 &51.3 \\ 
                    \midrule
                     &\textbf{Ours}$\dagger$  &64.2 & 49.8 & 26.9 & 52.2 & & 65.9 & 50.5 & 28.1 & 53.4  \\
                         &\textbf{Ours}     & \textbf{64.2} & \textbf{51.4} & \textbf{31.8} & \textbf{53.7} &  &\textbf{65.1} &\textbf{52.8} &\textbf{34.2} &\textbf{55.0}\\
                     \bottomrule
\end{tabular}
\label{imagenetlt_result}
\end{table}
\setlength{\tabcolsep}{1.4pt}

\subsection{Long-tailed Image Classification Task}
\label{section:Long-tailed Image Classification Task}
\subsubsection{Datasets and Setup.} 
We perform experiments on long-tailed image classification datasets, including the  ImageNet-LT~\cite{openlongtailrecognition} and iNaturalist2018~\cite{van2018inaturalist}.  
\begin{itemize}
        \item ImageNet-LT is a long-tailed version of the ImageNet dataset by sampling a subset following the Pareto distribution with power value 6. It contains 115.8K images from 1,000 categories, with class cardinality ranging from 5 to 1,280.
        \item iNaturalist2018 is the largest dataset for long-tailed visual recognition. It contains 437.5K images from 8,142 categories. It is extremely imbalanced with an imbalance factor of 512.
\end{itemize}

\par
\noindent
\subsubsection{Experimental Details.}   For image classification on ImageNet-LT, we implement all experiments in PyTorch. Following ~\cite{tang2020long,cui2021parametric,LADE}, we use ResNetXt-50~\cite{ResNext} as the feature extractor for all methods. We conduct model training with the SGD optimizer based on batch size 512, momentum 0.9. In both training epochs (90 and 200 training epochs), the learning rate is decayed by a cosine scheduler~\cite{sgdr}.  On iNaturalist2018~\cite{van2018inaturalist} dataset, we use ResNet-50~\cite{ResNext} as the feature extractor for all methods with 200 training epochs, with the same experimental parameters set for the other.  By default, learnable $\kappa$ for all categories are initialized as $16$ and $\lambda$ is $0.2$. Moreover, we use the same basic data augmentation (i.e., random resize and crop to 224, random horizontal flip, color jitter, and normalization) for all methods. 

\setlength{\tabcolsep}{14pt}
\begin{table}[t]
\begin{center}
\caption{
Benchmarking on iNaturalists2018 in accuracy (\%).  DT, CD and RL indicate decouple training, classifier design and representation learning, respectively. }

\label{table:inaturalists}
\begin{tabular}{l|l|cccc}
\toprule
\multirow{2}{*}{Type}  &\multirow{2}{*}{Method} & \multicolumn{4}{c}{iNaturalist2018}   \\ \cmidrule{3-6} &  & Many & Med. & Few &All  \\ \midrule
 \multirow{1}{*}{Baseline}      &CE              & 72.2 & 63.0  & 57.2 &61.7   \\ \midrule

\multirow{2}{*}{DT}   &Decoupling~\cite{2019Decoupling} &65.6 &65.3 &65.5&65.6\\
                    &BBN~\cite{zhou2020bbn}      & 49.4 & 70.8  & 65.3 &66.3 \\
                     \midrule
\multirow{2}{*}{CD}   &TDE~\cite{tang2020long}  &- &- &- &68.7\\

   &$\tau$-norm~\cite{2019Decoupling}                & 65.6 & 65.3  &65.5 &65.6 \\
                  \midrule
\multirow{2}{*}{RL}                     &TSC~\cite{li2022targeted}                   &72.6 &70.6   &67.8&69.7    \\
                    &DisAlign~\cite{zhang2021distribution} & 69.0 & 71.1   &70.2&70.6   \\ 
                     \midrule 
                     &\textbf{Ours}    & \textbf{72.8} & \textbf{71.7} & \textbf{70.0} & \textbf{71.0} \\
                     \bottomrule
\end{tabular}
\end{center}
\end{table}
\setlength{\tabcolsep}{1.4pt}

\par
\noindent
\subsubsection{Comparison with State of the Arts.}
In our paper, the comparison methods use single models. Note that there are also ensemble models for long-tailed classification, e.g., RIDE~\cite{RIDE} and TADE~\cite{TADE}. For fair comparisons, following xERM~\cite{xERM}, we will not include their results in the experiments. Tab.~\ref{imagenetlt_result} shows the long-tailed results on ImageNet-LT. We adopt the performance data from the deep long-tailed survey~\cite{longtailedsurvey} for various methods at 90 and 200 training epochs to make a fair comparison. Our approach achieves 53.7\% and 55.0\% in overall accuracy, which outperforms the state-of-the-art methods by a significant margin at 90 and 200 training epochs, respectively. Compared with representation learning methods, our method surpasses SSP by 0.6\% (53.7\% vs 53.1\%) at 90 training epochs and outperforms SSP by 1.7\% (55.0\% vs 53.3\%) at 200 training epochs. In addition, our method obtains higher performance by 1.0\% (53.7\% vs 52.7\%) and 0.6\% (55.0\% vs 54.4\%) comparing to PaCo at 90 and 200 training epochs, respectively. We observe that our vMF classifier (w/o proposed additional loss terms and post-training calibration algorithm) still achieves better performance than previous classifier design strategies, i.e., our vMF classifier surpasses $\tau$-norm and TDE which by 2.6\% (52.2\% vs 49.6\%) and 0.4\% (52.2\% vs 51.8\%) at 90 epochs. Moreover, our vMF classifier performs better when training 200 epochs than 90 epochs (53.4\% vs 52.2\%), in contrast to TDE (51.3\% vs 51.8\%). This shows that our vMF classifier has more potential to fit data better and learn better representations. 

Furthermore, Tab.~\ref{table:inaturalists} presents the experimental results on the naturally-skewed dataset iNaturalist2018. Compared with the improvement
brought by representation learning and classifier design approaches, our method achieves competitive result (71.0\%) consistently.

\subsection{Long-tailed Semantic and Instance Segmentation Task}
\label{section:Semantic Semgnetaion on ADE20k Dataset}
 To further validate our method, weconduct comprehensive experiments on the semantic and instance segmentation datasets, i.e., ADE20K~\cite{zhou2017scene} and LVIS-v1.0~\cite{gupta2019lvis}.
 \par
\noindent
 \textbf{Dataset and Setup. } 
 \begin{itemize}
        \item  ADE20K is a scene parsing dataset covering 150 fine-grained semantic concepts and it is one of the most challenging semantic segmentation datasets. The training set contains 20,210 images with 150 semantic classes. The validation and test set contain 2,000 and 3,352 images respectively.
        \item  LVIS-v1.0 contains 1230 categories with both bounding box and instance mask annotations.  LVIS-v1.0 divides all categories into 3 groups based on the number of images that contain those categories: frequent ($>$100 images), common (11-100 images) and rare ($<$10 images). We train the models with 57K train images and report the accuracy on 5K val images.
\end{itemize}

\setlength{\tabcolsep}{2.6pt}
\begin{table}[t]
\centering
\caption{Performance of semantic segmentation on ADE20K and instance segmentation on LVIS-v1.0. R-50 and R-101 denote ResNet-50 and ResNet-101, respectively. 'Cascade-R101' is for Cascade Mask R-CNN~\cite{cascade}.}
\begin{tabular}{l|l|cc|c|l|l|c|c}
\toprule
\multirow{2}{*}{Model} & \multirow{2}{*}{Method}   & \multicolumn{2}{c|}{ADE20K} & &\multirow{2}{*}{Model} &\multirow{2}{*}{Method} & \multicolumn{2}{c}{LVIS-v1.0}\\ \cmidrule{3-4} \cmidrule{8-9}  
               &                 & mIoU             & mAcc & & & &AP &AP$_b$         \\  \midrule
                          OCRNet & Baseline  &40.8       & 50.9 & &\multirow{8}{*}{\thead{Cascade \\(R101)}}  & Cross-Entropy  & 22.6 & 25.2        \\
                          (HRNet-W18)  & \textbf{Ours}                                    &  \textbf{41.5}   &  \textbf{52.9}       &      & &De-confound~\cite{tang2020long}   & 23.5    &25.8
                                   \\  \cline{1-4} 
                            
\multirow{3}{*}{\thead{DeepLabV3+ \\(R-50)}} & Baseline     & 44.9         & 55.0  &        &&TDE~\cite{tang2020long}    &27.1 &30.0     \\
                            & DisAlign~\cite{zhang2021distribution}                          & 45.7         &\textbf{57.3} &        & &EQL v2~\cite{EQL_V2}  &28.8 &32.3         \\
                            & \textbf{Ours}                                 & \textbf{ 45.9}  & 57.0  &      & & DisAlign~\cite{zhang2021distribution}  &28.9 &32.7  \\ \cline{1-4} 
\multirow{3}{*}{\thead{DeepLabV3+ \\(R-101)}}                          & Baseline   & 46.4         & 56.7 &       &  & BAGS~\cite{BAGS}    &27.9 &31.5      \\
                            & DisAlign~\cite{zhang2021distribution} & 47.1         & 59.5  &       &  & Seesaw Loss~\cite{seesaw}    &29.6 &32.5        \\
                            &\textbf{Ours}     &     \textbf{47.2}       &   \textbf{59.8}   && &   \textbf{Ours}    & \textbf{29.8} & \textbf{32.9} \\ \bottomrule
\end{tabular}
\label{ADE20K_result}
\end{table}
\setlength{\tabcolsep}{1.4pt}
\par
\noindent
\textbf{Experimental Details.} 
We evaluate our method using two wide-adopted segmentation models (OCRNet~\cite{ocrnet} and DeepLabV3+~\cite{deeplabv3+}) based on different backbone networks.  We initialize the backbones using the models pre-trained on ImageNet~\cite{imagenet} and the framework randomly. All models are trained with an image size of $512 \times 512$ and 160K iterations in total. We train the models using Adam optimizer with the initial learning rate 0.01, weight decay 0.0005 and momentum 0.9.   Furthermore, We implement our method on LVIS-v1.0 with mmdetection~\cite{mmdetection} and train Mask R-CNN~\cite{MASK_R-CNN} with random sampler by 2x training schedule. The model is trained with batch size of 16 for 24 epochs.  The optimizer is SGD with momentum 0.9 and weight decay 0.0001. The initial learning rate is 0.02 with 500 iterations’ warm up.  For above two tasks, we set the optimal configuration in our experiments that is all learnable $\bm{\mathcal{K}}$ are initialized to $16$.

\par
\noindent
\textbf{Comparison with State of the Arts.}
For the semantic segmentation task, The numerical results and comparison with other peer methods are reported in left part of Tab.~\ref{ADE20K_result}. Our method achieves 0.7\% (41.5\% vs 40.8\%) improvement in mIoU using OCRNet with HRNet-W18. Moreover, our method outperforms the baseline with large at  1.0\% (45.9\% vs 44.9\%) in mIoU using DeeplabV3+ with ResNet-50 when the iteration is 160K. Even with a stronger backbone: ResNet-101, our method also achieves 0.8\% (47.2\% vs 46.4\%) mIoU improvement than baseline. For the instance segmentation task, we report quantitative results and compare our method with recent work in the right part of Tab.~\ref{table:inaturalists}. Our method can achieve 29.8\% in AP and 32.9\% in $\rm AP_b$ when applied to the Cascade-R101. Apart from the CE loss baseline, we further compare our method with recent designs for long-tailed instance segmentation. Our method surpasses Seesaw Loss by 0.2\% (29.8\% vs 29.6\%) AP, and surpasses DisAlign by 0.9\% (29.8\% vs 28.9\%) AP, which reveals the effectiveness of our method.
\setlength{\tabcolsep}{12pt}
\begin{table}[t]
\centering
 \caption{Ablation on our proposed two loss terms and the loss weight $\lambda$. `None' indicates only the performance loss term is applied to train model. `$0.1$' means $\lambda$ is set as $0.1$. }
    \label{tab:loss}
    \centering
    \begin{tabular}{l|p{.8cm}<{\centering} p{.8cm}<{\centering} p{.8cm}<{\centering} p{.8cm}<{\centering}}
      \toprule
      Additional Loss & All & Many & Med. & Few \\
      \midrule
      Baseline & 51.8 & 62.6 & 48.9 & 31.3 \\
      None & 52.2 & 64.2 & 49.8 & 26.9 \\
      $0.2$, $\mathcal{L}_{icd}$ & 52.9 & 64.2 & 49.8 &  {31.7} \\
      $0.2$, $\mathcal{L}_{cfc}$ & 53.1 & 65.3 &  50.4 &  27.9\\
      $0.2$, $\mathcal{L}_{icd}$, $\mathcal{L}_{cfc}$ & \textbf{53.5} &  \textbf{65.4} & \textbf{50.8} &  \textbf{29.1}\\
      $0.1$, $\mathcal{L}_{icd}$, $\mathcal{L}_{cfc}$ & 53.2 &  65.0 & 50.7 &  27.9 \\
      $0.4$, $\mathcal{L}_{icd}$, $\mathcal{L}_{cfc}$ & 52.6 &  64.9 & 50.1 &  26.8 \\
      $0.3$, $\mathcal{L}_{icd}$, $\mathcal{L}_{cfc}$ & 53.2 &  65.1 & 50.8 &  28.0 \\
      \bottomrule
    \end{tabular}
\end{table}
\setlength{\tabcolsep}{1.4pt}

\setlength{\tabcolsep}{12pt}
\begin{table}[t]
\centering
 \caption{Ablation on the hyper-parameter $\alpha$ of post-training calibration algorithm with different classifiers. $\ddagger$ indicates the  corresponding classifier is calibrated under the optimal  $\alpha$. }
    \label{tab:kappa}
    \centering
    
    \begin{tabular}{l|p{.8cm}<{\centering} p{.8cm}<{\centering} p{.8cm}<{\centering} p{.8cm}<{\centering}}
      \toprule
      $\bm{\mathcal{K}}$ & All & Many & Med. & Few \\
      \midrule
      Linear & 43.2  &   66.2   &  35.4    &   6.0 \\
      Linear $\ddagger$ & 48.3 & 60.9 & 46.4  & 19.9 \\
      $\tau$-norm~\cite{2019Decoupling}  & 48.6 & 69.9 & 42.5 & 10.1 \\
      $\tau$-norm~\cite{2019Decoupling} $\ddagger$  & 53.0 & 66.5 & 50.2 & 24.1 \\
      Causal~\cite{tang2020long}  & 49.0 & 69.6 & 43.0 & 12.2 \\
      Causal~\cite{tang2020long} $\ddagger$ & 50.9 & 69.0 & 45.8  & 17.5  \\
      Ours  & 53.5 & 65.4 & 50.8 & 29.1 \\
      Ours $\ddagger$ & \textbf{53.7} & \textbf{63.9} & \textbf{51.5}  & \textbf{32.4}  \\
      \bottomrule
    \end{tabular}
\end{table}
\setlength{\tabcolsep}{1.4pt}

\subsection{Ablation Study} 
\label{section:ablation study}
We conduct ablation study on ImageNet-LT dataset to further understand the hyper-parameters of our methods and the effect of each proposed component. 

\subsubsection{Ablation study on two additional loss terms and the loss weight $\lambda$}
Firstly, we evaluate the effectiveness of the proposed $\mathcal{L}_{icd}$ and $\mathcal{L}_{cfc}$. Setting $\lambda=0.2$ and initializing $\kappa = 16$, we train vMF classifier w/o additional loss terms, w/ $\mathcal{L}_{icd}$, w/ $\mathcal{L}_{cfc}$ and w/ both of them, respectively. Experimental results are reported in Row 1-4 of Tab.~\ref{tab:loss}. Our baseline is the balanced cosine classifier~\cite{balanced_loss}.
Conclusions are \textbf{(1).} Giving additional surveillance via $\mathcal{L}_{icd}$ is beneficial to the performance on tail classes. It can be seen from the second and third rows in the Tab.~\ref{tab:loss}. The performance of the tail of the loss term has been greatly improved (26.9\% vs 31.7\%). \textbf{(2).} $\mathcal{L}_{cfc}$ gains the non-trival performance improvements on all classes. \textbf{(3).} Simultaneously adopting the above two loss terms further improves the accuracy by 1.3$\%$, further widening the performance gap up to 1.7$\%$ compared with the baseline. 
Secondly, we conduct four experiments on different $\lambda$. Row 5-8 of Tab.~\ref{tab:loss} show $\lambda=0.2$ is the optimal setting.

\subsubsection{Ablation study on post-calibration algorithm with different classifier}
To verify the versatility of our post-training calibration algorithm, we perform it on our vMF, linear, $\tau$-norm ($\tau=0.7$, optimal setting in~\cite{2019Decoupling}) and causal~\cite{tang2020long} classifiers, following Sec.~\ref{sec:calibrate_classifier}. All of them have trained on ImageNet-LT with ResNetXt-50. We set the hyper-parameter $\alpha$ in the interval 0 to 1 with a stride of 0.1 and take the eleven sets of values to conduct ablation experiments on above classifiers. For linear classifier, the optimal $\alpha=0.7$, where our algorithm improves allover accuracy performance by 5.1\%. For $\tau$-norm classifier and causal classifier, under the optimal $\alpha = 0.1$, the allover accuracy is improved by 4.4\% and 1.9\%. When $\alpha = 0.2$, our vMF classifier achieves highest accuracy 53.7\%.
The reason for slight improvement on ours may be because it has already learned with proposed loss terms ($\mathcal{L}_{cfc}$  and  $\mathcal{L}_{icd}$) that are also based on distribution overlap coefficient.

\section{Conclusions}
In this paper, we extend cosine-based classifiers as a vMF distribution mixture model on hyper-sphere, denoted as the vMF classifier. Benefiting from the representation space constructed by the vMF classifier, we define the distribution overlap coefficient to measure the representation quality for features and classifiers. Based on distribution overlap coefficient, we formulate the inter-class discrepancy and class-feature consistency loss terms to alleviate the interference among the classifier weights and align features with classifier weights. Furthermore, we develop a novel post-training calibration algorithm to zero-costly boost the performance. Our method outperforms previous work with a large margin and achieves state-of-the-art performance on long-tailed image classification, semantic segmentation, and instance segmentation tasks. 

\par 
\noindent
\textbf{Acknowledgments} This work is supported by the National Natural Science Foundation of China (U21B2004),  the Zhejiang Provincial key RD Program of China (2021C01119) , and the Zhejiang University-Angelalign Inc. R $\&$ D Center for Intelligent Healthcare.
\bibliographystyle{splncs04}
\bibliography{egbib}
\end{document}